\documentclass{article}

\usepackage[dblblindworkshop,final]{neurips_2026}

\usepackage[utf8]{inputenc} % allow utf-8 input
\usepackage[T1]{fontenc}    % use 8-bit T1 fonts
\usepackage{hyperref}       % hyperlinks
\usepackage{url}            % simple URL typesetting
\usepackage{booktabs}       % professional-quality tables
\usepackage{amsfonts}       % blackboard math symbols
\usepackage{nicefrac}       % compact symbols for 1/2, etc.
\usepackage{microtype}      % microtypography
\usepackage{xcolor}         % colors
\usepackage{amsmath,amssymb}
\usepackage{graphicx}
\usepackage[noabbrev,nameinlink,capitalize]{cleveref}

\definecolor{linkcolor}{RGB}{83,83,182}
\hypersetup{
    colorlinks=true,
    citecolor=linkcolor,
    linkcolor=linkcolor
}

\workshoptitle{TS-LIMITS}
\title{Investigating the effects of the patch size in Time Series Foundation Models}

\author{%
  Jad Yehya\thanks{Equal contribution.} \\
  Université Paris-Saclay, Inria, CEA \\
  Palaiseau, France \\
  \texttt{jad.yehya@inria.fr}
  \And
  Andrei-Tiberiu Pantea\footnotemark[1] \\
  CMAP, École polytechnique, IP Paris\\
  Électricité de France R\&D\\
  Palaiseau, France\\
  \texttt{andrei-tiberiu.pantea@polytechnique.edu}
  \And
  Aymeric Dieuleveut \\
  CMAP, École polytechnique, IP Paris \\
  Palaiseau, France
  \AND
  Matthieu Kowalski \\
  Université Paris-Saclay, Inria, CNRS, LISN \\
  Gif-sur-Yvette, France
  \And
  Thomas Moreau \\
  Université Paris-Saclay, Inria, CEA \\
  Palaiseau, France
}

\begin{document}

\maketitle

\begin{abstract}
Patch size is a central design choice in patch-based time-series foundation models, controlling temporal resolution, sequence length, and computational cost. We study how this choice interacts with transformer architecture through a controlled comparison of encoder-only and decoder-only forecasting models trained with the same data mixture and optimization budget. Across 97 GIFT-Eval tasks, encoder-only models consistently benefit from smaller patches, whereas decoder-only performance follows a U-shaped trend, with an intermediate patch size performing best in aggregate. Dataset-level results further show that the preferred decoder patch size increases with the forecasting horizon. For encoder-only models, increasing context length improves performance across patch sizes, although the gains diminish beyond a context length of 2048. Finally, training-compute and CPU-latency Pareto analyses reveal complementary regimes: decoder-only models provide efficient low-cost alternatives, while encoder-only models attain the highest overall accuracy. These results show that the choice of patch size and architecture should be made depending on the forecasting horizon and deployment constraints.
\end{abstract}

%encoder-only~\citep{chronos2, Kairos, MOMENT, PatchTST-FM} or decoder-only~\citep{timesfm,moirai2,toto2,timer}.

\section{Introduction} \label{sec:sec01}
Recent years have witnessed an increasing interest in Time Series Foundation Models (TSFMs), which promise to replace dataset-specific forecasting pipelines with pretrained models capable of transferring across domains, sampling frequencies, context lengths, and prediction horizons, and, in some cases, across time series tasks.
The growing capabilities of TSFMs are reflected in their increasingly strong performance on the various general purpose time series forecasting benchmarks such as GIFT-Eval \citep{aksu2024giftevalbenchmarkgeneraltime}, fev-bench \citep{shchur2026fevbenchrealisticbenchmarktime}, TIME \citep{qiao2026itstimegenerationtime}.

Despite rapid empirical progress, the field has not converged towards a single architectural recipe.
The current state-of-the-art spans a wide range of backbones, including the Tirex family~\citep{auer2025tirex} of recurrent models built around xLSTM layers and state-space-based FlowState~\citep{graf2026flowstate}.
Significant diversity also exists in the class of patch-based Transformers~\citep{vaswani2017attention}: PatchTST-FM~\citep{Wen2026RevisitingTG} and Chronos-2~\citep{ansari2025chronos2} use encoder-only, full-attention backbones to predict masked future patches in parallel, while TimesFM~\citep{das24timesfm}, Moirai 2.0~\citep{liu2026moirai20timeseries}, and Toto~\citep{cohen2025toto} employ causal decoder-only transformer layers for single or multiple next-patches forecasting.

This diversity among the state-of-the-art shows that benchmark performance alone has not, until now, identified a preferred architecture for TSFMs.
Moreover, published models differ simultaneously in the pretraining corpus, scale, training objectives, and tokenization strategies, preventing the isolation of the effects due to each component in the architectures.

Among these design choices, patch size remains insufficiently studied. 
Existing ablations of patch sizes have been performed within individual model families and evaluated through aggregate forecasting error metrics on benchmarks, e.g. GIFT-Eval. For example, \citet{Yuqietal-2023-PatchTST} report relatively robust PatchTST performance across patch lengths, with 8 and 16 generally performing well, while \citet{das24timesfm} study input patch length in TimesFM and decouple input and output patching, using lengths 32 and 128, respectively. 
Notably, patch size has also been connected to the rank structure and compressibility of TSFM representations~\citep{yu2026understanding}. 
However, these analyses remain architecture-specific, and the effect of patch size has not been systematically compared between encoder-only and decoder-only TSFMs under matched training conditions.

\begin{figure}[t]
    \centering
    % Replace with the original filename if you still have the figure source.
    \includegraphics[width=\linewidth]{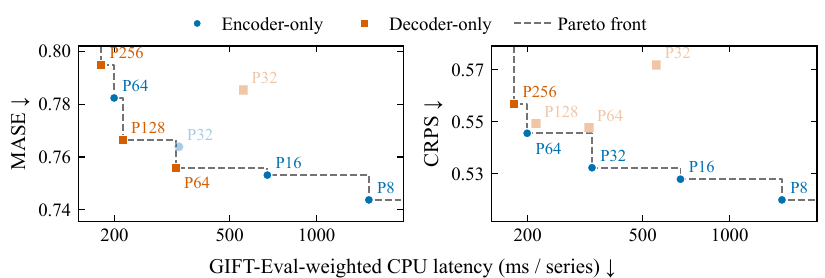}
    \caption{Normalized MASE (left) and CRPS (right) against CPU latency for encoder-only and decoder-only models for various values of the patch size $P$. The latency is computed over evaluations on the GIFT-Eval horizon lengths and averaged according to GIFT-Eval's proportions.}
    \label{fig:latency}
\end{figure}

To address this gap, we conduct a controlled study of patch size in encoder-only and decoder-only patch-transformer-based TSFMs.
We train both architectures under matched conditions and, within each architecture, vary the patch size while maintaining the remaining experimental setup fixed. This allows us to isolate the effect of temporal tokenization and determine whether patch size interacts differently with full-attention and causal-attention backbones.
\cref{fig:latency} previews the central accuracy-efficiency trade-off identified in our experiments. Decoder-only configurations dominate the low-latency regime, while encoder-only models extend the Pareto frontier towards higher forecasting accuracy.
The position of each configuration changes substantially with patch size, motivating a joint analysis of tokenization, architecture, and computational cost.
The paper is structured as follows:

\begin{itemize}
    \item In \cref{sec:models}, we introduce the encoder-only and decoder-only architectures studied in this work and explain their particularities. Both follow closely state-of-the-art architectures for encoder-only and decoder-only TSFMs.
    \item In \cref{sec:experiments}, we begin with the experimental setup and then present the numerical results of our experiments. We provide per family fine grained evaluations on GIFT-Eval~\citep{aksu2024giftevalbenchmarkgeneraltime} and aggregated performance-efficiency comparisons between encoder-only and decoder-only models.
    \item In \cref{sec:conclusion}, we comment on the limitations and present the conclusions of our work.
\end{itemize}

\section{Models} \label{sec:models}

We study the effect of patch size in two patch-based TSFM configurations: an encoder-only model based on PatchTST-FM~\citep{Wen2026RevisitingTG} and a causal decoder-only family following the autoregressive patch-decoder paradigm used by TimesFM~\citep{das24timesfm} and Toto~\citep{cohen2025toto}.
More specifically, the decoder follows the architecture of Toto, retaining its causal normalization, linear patch embeddings, next-patch training, and transformer architecture, while replacing its distributional output head with a monotonic quantile head.

Both models operate on non-overlapping patches and produce probabilistic forecasts through quantile regression.
They differ in their attention pattern and prediction objective: the encoder uses full attention to reconstruct masked patches, whereas the decoder uses causal attention to predict future patches.
Within each model family, keeping the remaining components fixed allows us to study how the effect of patch size depends on the underlying architecture.

\paragraph{Probabilistic forecasting setting}

The problem we study is that of univariate time series probabilistic forecasting.
Let $x_{1:C}=(x_1,x_2,\ldots,x_C)\in \mathbb{R}^C$ denote an observed context of length $C$ and let $x_{C+1:C+H}$ denote the subsequent prediction horizon of length $H$.
Given the context, the model predicts $Q$ conditional quantiles for every horizon time stamp. For a set of quantile levels $\mathcal{Q}=\{\tau_1,\tau_2,\ldots,\tau_Q\}\subset(0,1)$, the model's output is therefore $\hat{y}\in\mathbb{R}^{H\times Q}$ and the goal is to have
\begin{equation}
\hat{y}_{h,k} \approx F^{-1}_{x_{C+h}\mid x_{1:C}}(\tau_k),
\qquad h\in\{1,2,\ldots,H\},\quad k\in\{1,2,\ldots,Q\}.
\label{eq:quantiles}
\end{equation}

\paragraph{Patching}
Both model families divide the input into contiguous, non-overlapping patches of size $P$. If the context length is not divisible by $P$, the context is zero-padded on the left so that one obtains $N_C \colon = \lceil\nicefrac{C}{P}\rceil$ patches.

Forecasting a horizon of length $H$ then requires $N_H \colon = \lceil\nicefrac{H}{P}\rceil$ future patches. 
When $H$ is not divisible by $P$, the final predicted patch is truncated to retain exactly the requested $H$ time steps.

\subsection{Encoder-only model}
\label{sec:encoder}

The encoder-only family is trained as a masked reconstruction model: selected patches are hidden from the input, and their values are predicted simultaneously using full self-attention.

\paragraph{Masking and normalization.}

We use contiguous-patch masking (CPM), following \citet{auer2025tirex}, to mask contiguous spans inside the context and at the end, for forecasting.
Our only departure from PatchTST-FM is to mask contiguous sequences of $128$ time points inside the context rather than a fixed number of patches, so that the masked span, and hence the reconstruction problem seen during training, is identical for every patch size.
We retain the binary mask $m\in\{0,1\}^C$ where $m_t=1$ if $x_t$ is neither masked nor padded and 0 otherwise.

We then normalize the input time series using Mask-Aware Normalization introduced in \citep{Wen2026RevisitingTG}. 
That means that time series specific mean $\mu$ and standard deviation $\sigma$ are computed only using the visible time points (i.e. where $\mathbf{m} = 1$).
The same statistics are used to transform the corresponding predictions back to the original scale.

Furthermore, the influence of outliers is reduced using $\text{arcsinh}$, as introduced by \citep{ansari2025chronos2}. 
The normalized series then becomes
\begin{equation} \label{eq:normalization}
    \bar{\mathbf{x}} = \sinh^{-1}\left(\frac{\mathbf{x} - \mu}{\sigma}\right).
\end{equation}

\paragraph{Patch embedding and backbone.}

For every patch, its normalized values are concatenated with the corresponding binary mask. 
A residual block maps the resulting vector to the encoder embedding dimension $d$, to which learnable positional embeddings are added.
The tokenization procedure thus produces a sequence of patch embeddings $\mathbf{e} \in \mathbb{R}^{N_C \times d}$ that are further processed by the transformer layers.

\paragraph{Prediction.}

The models append $N_H \colon = \lceil\nicefrac{H}{P}\rceil$ masked patches representing the prediction horizon to the observed context patches.
The hidden representation associated with each masked patch is processed by a residual output head that produces $P \times Q$ values which  represent $Q$ conditional quantiles for each of the $P$ time points in the patch.
Since the encoder processes all context and masked tokens jointly, the complete horizon is reconstructed in a single forward pass.

\subsection{Decoder-only model} \label{sec:sec02c}
The decoder-only family is trained autoregressively: each patch is predicted from all preceding patches using causal self-attention.

\paragraph{Normalization.}
We normalize each patch using causal, prefix-dependent statistics.
For the $i$-th patch, the mean $\mu_i$ and standard deviation $\sigma_i$ are computed using only the valid observations up to the end of that patch. 
The normalized patch is therefore
\begin{equation} \label{eq:decoder_normalization}
\bar{\mathbf{p}}_i = \frac{\mathbf{p}_i-\mu_i}{\sigma_i}.
\end{equation}
The same statistics are used to transform the corresponding predictions back to the original scale. 
Since patch $i$ is used to predict patch $i+1$, the normalization statistics never depend on the target values.

\begin{table}[t]
    \centering
    \caption{Configurations considered in our experiments. All the configurations are trained for 100k optimization steps.}
    \label{tab:configs}
    \begin{tabular}{lcc}
        \toprule
        Model family & Patch size $P$ & Context length $C$ \\
        \midrule
        Encoder-only & 4 & 1024 \\
        Encoder-only & 8 & 1024, 2048 \\
        Encoder-only & 16, 32, 64 & 1024, 2048, 4096 \\
        Decoder-only & 16, 32, 64, 128, 256 & 2048 \\
        \bottomrule
    \end{tabular}
\end{table}

\paragraph{Patch embedding and backbone.}

For every patch, its normalized values are concatenated with the corresponding binary visibility mask.
A bias-free linear projection maps the resulting vector to the decoder embedding dimension $d$.
The tokenization procedure thus produces a sequence of patch embeddings $\mathbf{e}\in\mathbb{R}^{N_C\times d}$.

The patch embeddings are processed by a stack of causal transformer layers. 
No positional embedding is added, positional information is introduced only through RoPE \citep{su2024roformer} inside the transformer layers.

\paragraph{Prediction.}

The prediction is performed autoregressively, patch by patch.
The hidden representation at each position is passed through a linear output projection producing $P\times Q$ values.
The central output directly represents the median, while the remaining outputs are transformed into positive increments using $\operatorname{softplus}$.
Cumulative increments are subtracted from or added to the median to obtain the lower and upper quantiles, respectively.
The predicted quantiles are therefore ordered by construction.

\section{Experiments} \label{sec:experiments}

The goal of our experiments is to investigate how patch size affects forecasting accuracy, sensitivity to the prediction horizon, and training efficiency.
We first study patch size at a fixed context length within each model family, but varying the patch size while keeping the context and prediction lengths fixed is not enough for a full picture.
To that end, for the encoder-only family, we additionally vary the context length to examine the interaction between temporal resolution and context sequence length.
Finally, we compare the pretrained configurations in terms of forecasting performance across multiple horizon lengths.

\subsection{Experimental Setup} \label{sec:methodo}

\paragraph{Model configurations.}

\cref{tab:configs} summarizes the configurations considered in our experiments. 
All the models are trained for 100k optimization steps. 
Within each sweep, patch size is varied while the remaining model and training configuration is held fixed.
The encoder experiments cover context lengths between $1024$ and $4096$, whereas the decoder is trained at a context length of $2048$.
Each configuration is pretrained with multiple random seeds that were allocated sequentially: configurations that clearly underperformed in the initial runs were not replicated to the maximum number, allowing the available compute to be concentrated on competitive configurations and on estimating variability around the observed optima.
All seed-level aggregates use the available runs for each configuration, whose numbers are reported in \cref{tab:seeds}.
More details about the pretraining procedure and full model configurations are in \cref{app:architectures}.

\paragraph{Evaluation protocol.}

All the configurations are evaluated on the 97 forecasting tasks of GIFT-Eval~\citep{aksu2024giftevalbenchmarkgeneraltime}.
We report normalized MASE (mean absolute scaled error) to evaluate point forecasts and normalized CRPS (continuous ranked probability score) to evaluate probabilistic forecasts.
The normalization is performed with respect to the performance of the seasonal naive predictor and aggregated results are obtained by geometrically averaging the normalized task-level scores, following the approach in~\citep{aksu2024giftevalbenchmarkgeneraltime}.

We also report finer grained results, namely the dataset win rates of each patch size, defined as the fraction of tasks on which it obtains the lowest error among the configurations in the corresponding sweep. 
Furthermore, we also present these win rates according to the short, medium, and long-term forecasting categories of GIFT-Eval, which contain 55, 21, and 21 tasks, respectively.  

%\paragraph{Training efficiency.}

%Equalizing the number of optimization steps does not equalize training cost. For a fixed context length, decreasing $P$ increases the number of input tokens from approximately $C/P$, thereby increasing the computational cost of self-attention and the remaining token-wise operations. We consequently measure the A100 GPU-hours required to complete 100k training steps and analyze the resulting accuracy-compute trade-off.

\subsection{Encoder-only patch size sweep}
\label{sec:encoder_results}

\begin{figure}[h!]
    \centering
    \includegraphics[width=0.49\textwidth]{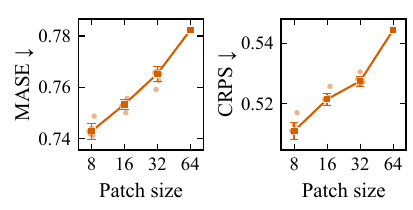}
    \hfill
    \includegraphics[width=0.49\textwidth]{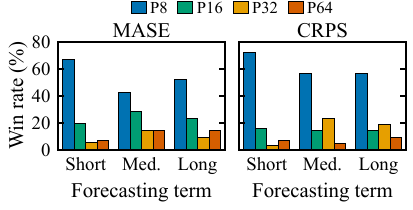}
    \caption{Effect of patch size on the encoder-only model for $C = 2048$. \textbf{Left:} normalized MASE and CRPS on GIFT-Eval. \textbf{Right:} dataset win rates by forecasting term. The encoder-only model with patch size $P = 8$ is the best in both aggregate performance and dataset win rates within each term.}
    \label{fig:encoder_patch_sweep}
\end{figure}

We first compare encoder-only models with patch sizes $P \in \{8, 16, 32, 64\}$ at a fixed context length $C = 2048$.
\cref{fig:encoder_patch_sweep} shows a consistent deterioration in both normalized MASE and normalized CRPS as the patch size increases.
Among the evaluated configurations, $P=8$ therefore provides the most accurate point and probabilistic forecasts.

The dataset-level results show that this improvement is broadly distributed across GIFT-Eval rather than being driven by a small number of datasets.
The $P=8$ model obtains the best MASE on 57 of the 97 tasks, and the best CRPS on 64 tasks.
By comparison, $P=16$ wins on 22 and 15 tasks, respectively, while $P=32$ wins on 8 and 11 tasks and $P=64$ wins on 10 and 7 tasks.

The benefit of smaller patches is most pronounced for short-term forecasting, but persists across medium and long horizons.
At fixed context length, smaller patches also yield longer token sequences.
This suggests that finer tokenization may improve performance by preserving more local information while allowing full attention to integrate it across the entire context.

%The term-wise analysis in Appendix~\ref{sec:additional_results} leads to a similar conclusion. The advantage of $P=8$ is particularly pronounced on short-term forecasts. For long-term MASE, $P=16$ becomes competitive with $P=8$, but $P=8$ remains stronger for CRPS across the three forecasting terms. Thus, over the range considered in this controlled sweep, the encoder consistently benefits from a finer temporal tokenization.

\subsection{Interaction Between Patch Size and Context Length} \label{sec:context_results}

\begin{figure}[h!]
    \centering
    \includegraphics[width=0.95\textwidth]{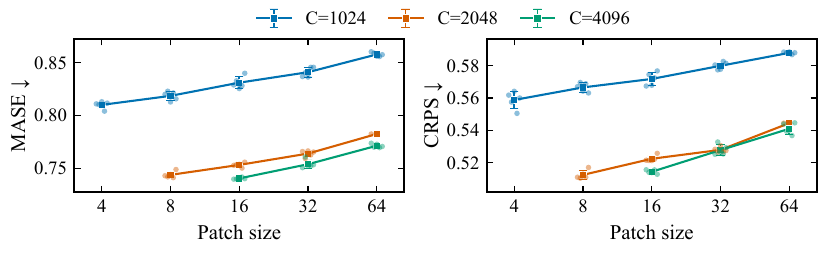}
    \caption{Encoder-only context and patch-size sweep after 100k training steps. Within each context length, smaller patches perform better. Increasing the context length substantially improves performance, although the gains from $2048$ to $4096$ are smaller than those from $1024$ to $2048$.}
    \label{fig:context_patch_sweep}
\end{figure}

Patch size does not only determine the temporal resolution of each token but also the number of tokens available to represent a given context.
Therefore, we extend the encoder-only analysis to context lengths $C\in\{1024,2048,4096\}$, as shown in \cref{fig:context_patch_sweep}.
This experiment complements \cref{sec:encoder_results}, as it appears that performance is not only related to the number of tokens per context, but also to the context length or patch size.

At every fixed context length, smaller patches produce lower MASE and CRPS.
Nonetheless, increasing the context length provides a substantial improvement across patch sizes.

In particular, the improvement from $C=1024$ to $C=2048$ is very important for all the patch sizes considered.
The gains from increasing the context from 2048 to 4096 are smaller and suggest a saturation of the pretraining pipeline used for the chosen configurations.

\subsection{Decoder-Only Patch-Size Sweep} \label{sec:decoder_results}

\begin{figure}[h]
    \centering
    \includegraphics[width=0.49\textwidth]{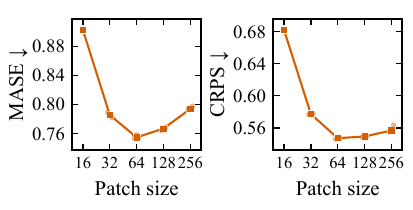}
    \hfill
    \includegraphics[width=0.49\textwidth]{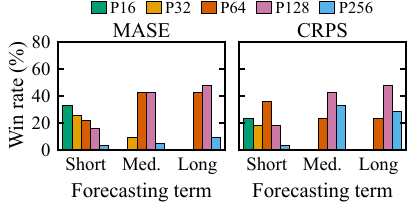}
    \caption{Effect of patch size on the decoder-only model. \textbf{Left:} aggregate MASE and CRPS on GIFT-Eval. \textbf{Right:} dataset win rates separated by forecasting term. Aggregate performance is best at $P = 64$, while the patch sizes favored by individual tasks shift as the forecasting term increases.}
    \label{fig:decoder_patch_sweep}
\end{figure}

The decoder-only models exhibit a different dependence on patch size.
As shown in \cref{fig:decoder_patch_sweep}, their aggregate performance follows a U-shaped curve rather than improving monotonically as the patch size becomes smaller.
The best overall results are obtained with $P = 64$ and both smaller and larger patches degrade performance.

The preferred patch size varies with the forecasting horizon, with larger patches becoming more competitive as the horizon increases.
For short-term tasks, $P=16$ achieves the most MASE wins (18/55), while $P=64$ achieves the most CRPS wins (20/55).
In contrast, $P=128$ performs best on medium- and long-term tasks, obtaining 9/21 and 10/21 MASE and CRPS wins, respectively.

These results explain why the intermediate patch size $P=64$ is the best across aggregate metrics: it is a compromise between the smaller patches favored by short-term tasks and the larger patches favored at longer horizons.
The aggregate optimum thus partly reflects the horizon distribution of GIFT-Eval, and a decoder-only model targeting a narrower horizon range would prefer a different patch size.

\subsection{Accuracy-Compute Trade-Off} \label{sec:compute_results}

\begin{figure}[t]
    \centering
    \includegraphics[width=0.99\textwidth]{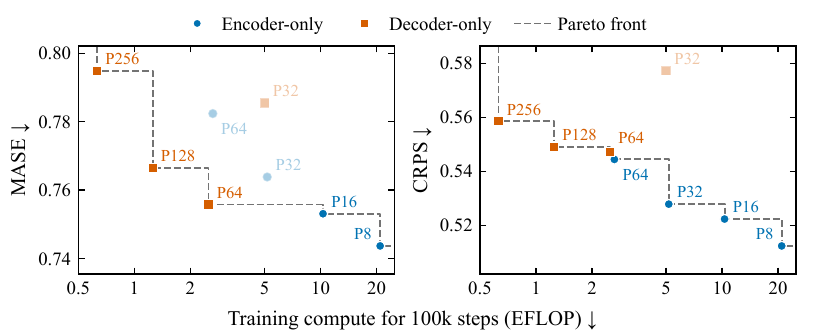}
    \caption{Normalized MASE (left) and CRPS (right) against training compute for encoder-only and decoder-only models with various patch sizes $P$. Training compute denotes the number of matrix-multiplication FLOPs required to train each model for 100k steps with an effective batch size of 512 and is reported in EFLOP. Each point shows the mean performance over pretraining seeds and the dashed line indicates the Pareto front.
    }
    \label{fig:compute_pareto}
\end{figure}

\cref{fig:compute_pareto} plots forecasting performance against the number of EFLOPs required for 100k training steps of different configurations with a fixed context length of 2048. 
Patch size has a large effect on training cost because smaller patches produce longer token sequences. 

The Pareto frontier contains configurations for which no other evaluated model is both more accurate and less computationally expensive at training. Decoder-only models occupy the low-compute region of the frontier, with $P=128$ and $P=256$ providing the cheapest alternatives, while $P=64$ offers the best decoder-only accuracy. Encoder-only models extend the frontier towards higher accuracy, with $P=8$ achieving the lowest MASE and CRPS overall. However, moving from $P=16$ to $P=8$ doubles the training cost for a comparatively modest improvement in accuracy, indicating diminishing returns.

\section{Conclusions and limitations} \label{sec:conclusion}

We presented a controlled study of how patch size affects encoder-only and decoder-only patch-based TSFMs. Despite their common use of temporal patches, the two architectures exhibit qualitatively different behavior. Encoder-only models consistently benefit from finer patches, while decoder-only models achieve their best aggregate performance at an intermediate patch size. Moreover, the preferred decoder patch size increases with the forecasting horizon, showing that the aggregate optimum partly reflects the horizon distribution of the evaluation benchmark.

The context-length experiments further show that patch size and context should not be considered independently. 
Longer contexts improve encoder-only performance across patch sizes, although the gains diminish beyond $C=2048$ in our setting. 
The efficiency results reveal complementary operating regimes: decoder-only models occupy the low-training-compute and low-latency portions of the Pareto frontiers, whereas encoder-only models extend the frontiers towards higher accuracy. 
In particular, the best decoder configuration remains competitive while requiring approximately one eight of the training compute of the most accurate encoder configuration. 
Consequently, neither architecture nor patch size is universally optimal and the appropriate configuration depends on the desired balance between forecasting accuracy, training cost, inference latency, and prediction horizon.

As a practical rule within the evaluated model scale, $P=8$ is the preferred encoder-only configuration when accuracy is the priority, while decoder-only $P=64$ provides a strong compromise between accuracy and computational cost. For decoder-only models targeting primarily medium or long horizons, $P=128$ may be preferable, and $C=2048$ provides a practical context-length compromise for encoder-only models given the diminishing gains beyond it.

\paragraph{Limitations.}
Our conclusions are restricted to the architectures, model scales, data mixture, and training paradigm (100k-step training budget, pretraining corpus, etc.) considered here.
All models are trained for a fixed budget of 100k optimization steps and longer training or a compute-matched training budget could change the relative performance of different patch sizes, since patch size affects both sequence length and the amount of computation performed per step.
The two model families necessarily differ in components beyond their attention patterns, including normalization, positional encoding, output heads, and parameter counts; cross-family differences therefore cannot be attributed exclusively to the encoder-decoder distinction. 

In the decoder-only experiments, the same patch size is used for both inputs and outputs. 
Varying $P$ therefore changes both the input granularity and the number of autoregressive decoding steps, and the observed trends may not hold when input and output patch sizes are selected independently, as in TimesFM~\citep{das24timesfm}.

The evaluation is also limited to GIFT-Eval, whose distribution of datasets and forecasting horizons influences the reported aggregate optima. 
Finally, inference latency is measured at batch size one on a single laptop CPU and the resulting Pareto frontier may vary across hardware, numerical precision, software implementations, and batching regimes.

\bibliography{references}
\bibliographystyle{abbrvnat}

%%%%%%%%%%%%%%%%%%%%%%%%%%%%%%%%%%%%%%%%%%%%%%%%%%%%%%%%%%%%

\appendix

\section{Model architectures and training details}
\label{app:architectures}

This section provides additional implementation details for the two model families described in \cref{sec:models}. The encoder-only model is an implementation of PatchTST-FM~\citep{Wen2026RevisitingTG}. Its only modification to the original training setup is the use of a fixed number of masked time points rather than a fixed masking percentage. The decoder-only model follows the architectures and autoregressive training procedures of Toto~\citep{cohen2025toto} and TimesFM~\citep{das24timesfm}.

\subsection{Encoder-Only PatchTST-FM}

\paragraph{Fixed-length contiguous masking.}
The encoder-only model is trained using contiguous-patch masking. 
The context window of $C$ time points is divided into non-overlapping, patch-aligned spans, and spans are masked uniformly at random until about 40\% of the window is hidden. 
In addition, the last $\ell \sim \mathcal{U}\left(0, 4S\right)$ points are masked, where $S$ is the span length, so that part of the objective is forecasting. 
Our only change to PatchTST-FM is how the span length is set. 
PatchTST-FM fixes it in patches ($N_{\text{cpm}} = 8$), so the masked span would grow with patch size $P$. 
We instead fix it in time points, $S = 128$, i.e., $N_{\text{cpm}} = 128/P$. With this choice, the number of masked spans, their length, the length of the tail mask, and hence the reconstruction problem seen during training are the same for every patch size.

Let $\mathbf{m}\in\{0,1\}^{C}$ denote the visibility mask, with $\mathbf{m}_t=1$ for visible, non-padded observations and $\mathbf{m}_t=0$ for masked or padded positions. The time series mean and standard deviation are computed only from visible observations:
\begin{equation}
    \mu =
    \frac{\sum_{t=1}^{C}m_t x_t}{\sum_{t=1}^{C}m_t},
    \qquad
    \sigma =
    \sqrt{
        \frac{\sum_{t=1}^{C}m_t(x_t-\mu)^2}
             {\sum_{t=1}^{C}m_t}
    }.
\end{equation}
The visible values are then normalized and transformed using $\operatorname{arcsinh}$:
\begin{equation}
    \bar{x}_t =
    \operatorname{arcsinh}\left(\frac{x_t-\mu}{\sigma}\right).
\end{equation}
Masked and padded positions are set to zero after normalization.

\paragraph{Embedding and Transformer backbone.}
Each normalized patch $\bar{p}_i\in\mathbb{R}^P$ is concatenated with its patch-level visibility mask $m_i\in\{0,1\}^P$. A residual embedding block maps the resulting $2P$-dimensional vector to the model dimension $d$:
\begin{equation}
e_i = f_{\mathrm{enc}}\!\left(\begin{bmatrix}\bar{p}_i\\m_i\end{bmatrix}\right)+r_i,
\label{eq:encoder_embedding}
\end{equation}
where $r_i$ is a learned positional embedding. The complete sequence is processed using full bidirectional self-attention.

\paragraph{Prediction.}
During training, the output head reconstructs the masked patches, and the loss is evaluated only at masked, non-padded time points. At inference time, $N_H=\lceil H/P\rceil$ masked patches are appended to the observed context. Full attention jointly processes the observed and masked patches, allowing the complete forecasting horizon to be predicted in a single forward pass. The final patch is truncated when $H$ is not divisible by $P$.

\subsection{Decoder-Only Model}

\paragraph{Causal normalization.}
For each patch position $i$, the model computes normalization statistics from all observations up to the end of that patch:
\begin{equation}
\mu_i=\frac{1}{iP}\sum_{t=1}^{iP}x_t,
\qquad
\sigma_i=\sqrt{\frac{1}{iP}\sum_{t=1}^{iP}(x_t-\mu_i)^2}.
\label{eq:causal_stats}
\end{equation}

The observations in patch $i$ are normalized as
\begin{equation}
\bar{x}_t=\frac{x_t-\mu_i}{\sigma_i},
\qquad (i-1)P<t\le iP.
\label{eq:causal_norm_detail}
\end{equation}

The same statistics are used to transform the prediction of the next patch back to the original scale:
\begin{equation}
\hat{x}_t=\mu_i+\sigma_i\hat{\bar{x}}_t,
\qquad iP<t\le(i+1)P.
\label{eq:causal_denorm}
\end{equation}
Thus, each prediction uses only statistics computed from observations available before the predicted patch.

\paragraph{Embedding and causal backbone.}
The normalized values are concatenated with their binary visibility mask and projected to the model dimension using a bias-free linear layer:
\begin{equation}
e_i=W_{\mathrm{in}}
\begin{bmatrix}
\bar{p}_i\\m_i
\end{bmatrix},
\qquad W_{\mathrm{in}}\in\mathbb{R}^{d\times 2P}.
\label{eq:decoder_embedding}
\end{equation}

The embeddings are processed by a causal Transformer composed of RMSNorm, grouped-query self-attention, and SwiGLU feed-forward blocks. No additive positional embeddings are used. Patch positions are represented exclusively through rotary positional embeddings applied to the query and key representations inside the attention layers.

The causal attention mask ensures that the representation used to predict patch $p_j$ can depend on all preceding patches $p_1,\ldots,p_{j-1}$, but not on $p_j$ or any later patch. During training, predictions for all target positions are computed in parallel using teacher forcing.

\paragraph{Monotonic quantile head.}
For each predicted patch, a linear output projection produces $P\times Q$ unconstrained values. The central output directly parameterizes the median. The remaining outputs are transformed into positive increments using softplus. Starting from the median, cumulative increments are subtracted to construct lower quantiles and added to construct upper quantiles. Consequently, for ordered levels $\tau_1<\cdots<\tau_Q$, the predictions satisfy
\begin{equation}
\hat{y}_{t,\tau_1}\le\hat{y}_{t,\tau_2}\le\cdots\le\hat{y}_{t,\tau_Q}
\label{eq:ordered_quantiles}
\end{equation}
by construction.

At inference time, the complete observed context is first processed to predict one future patch. The predicted median is appended to the context and used when generating the following patch. Causal normalization is recomputed over the extended context, and generation continues until $N_H$ future patches have been produced.

\subsection{Quantile Training Objective}

Both model families are trained using the pinball loss. For a target $x_t$, quantile level $\tau$, and prediction $\hat{y}_{t,\tau}$, the loss is
\begin{equation}
\rho_\tau(x_t-\hat{y}_{t,\tau})
=\max\left\{\tau(x_t-\hat{y}_{t,\tau}),\,(\tau-1)(x_t-\hat{y}_{t,\tau})\right\}.
\label{eq:pinball}
\end{equation}
The objective is averaged across all quantile levels and prediction positions.

\subsection{Training Details}

The pretraining corpus consists of GIFT-Eval Pretrain, 1 million KernelSynth~\citep{ansari2024chronos} and synthetically generated time series using our RandomPatterns pipeline. The RandomPatterns pipeline works as follows. In order to generate a time series of length $L$:
\begin{enumerate}
    \item Sample an integer period length $p$ between 4 and 256;
    \item Sample a random trend type among linear, quadratic, periodic and their corresponding parameters;
    \item Generate a standard random walk $r$ of length $p$;
    \item Generate a trend time series $t$ of length $L$;
    \item The generated RandomPatterns time series is $x_i=t_i+r_{i\bmod p}$.
\end{enumerate}

This pipeline provides rougher fully predictable time series with random seasonalities, preventing the model from overfitting to dataset specific periods and also encouraging it to learn broader representations.

The dataset mixture is: 10\% GIFT-Eval, 50\% RandomPatterns, and 40\% KernelSynth. Each configuration is trained from scratch for 100k optimization steps using AdamW~\citep{loshchilov2018decoupled} with an effective batch size of 512. We use a peak learning rate of $1\mathrm{e}{-4}$ and the OneCycleLR scheduler.

\begin{table}[h!]
    \centering
    \caption{Architecture and parameter counts of the evaluated model
    families. Parameter counts vary slightly with patch size and context
    length; ranges are reported over all evaluated configurations.}
    \vspace{5pt}
    \label{tab:architecture}
    \begin{tabular}{lccccc}
        \toprule
        Model & $d_{\mathrm{model}}$ & Layers & Heads
              & Quantiles & Parameters \\
        \midrule
        Encoder-only & 1024 & 20 & 16 & 99
                     & $255.0$-$267.5$M \\
        Decoder-only & 1024 & 20 & 16 & 9
                     & $253.4$–$255.9$M  \\
        \bottomrule
    \end{tabular}
\end{table}

\cref{tab:architecture} summarizes the size of the two model families. The encoder-only configurations contain approximately 255-267 million parameters, while the decoder-only configurations contain approximately 253-256 million.
Note that the difference in the number of quantiles does not affect the evaluations since GIFT-Eval performances are computed on 9 quantiles (0.1 - 0.9) and classification is performed using the frozen layers before the quantile head.

\paragraph{Random seeds.}
To account for training variability, each configuration is pretrained with several random seeds.
\cref{tab:seeds} lists the seeds used for every configuration.

\begin{table}[h]
    \centering
    \caption{Number of pretraining seeds per configuration.}
    \vspace{5pt}
    \begin{tabular}{lcccccccc}
    \toprule
    \textbf{Model family} & \textbf{Context length} & $P=4$ & $8$ & $16$ & $32$ & $64$ & $128$ & $256$ \\
    \midrule
    Encoder-only & 1024 & 5 & 5 & 5 & 5 & 5 & -- & -- \\
    Encoder-only & 2048 & -- & 5 & 5 & 5 & 1 & -- & -- \\
    Encoder-only & 4096 & -- & -- & 5 & 5 & 5 & -- & -- \\
    \midrule
    Decoder-only & 2048 & -- & -- & 1 & 2 & 3 & 3 & 3 \\
    \bottomrule
    \end{tabular}
    \label{tab:seeds}
\end{table}

\subsection{Latency experiment}
\label{app:latency_exp}

CPU latency in \cref{fig:latency} was measured independently on an Apple M1 Pro MacBook Pro using batch size 1, four CPU threads, one warm-up iteration, and one timed iteration.
The latency benchmark reproduced the 13 distinct context–horizon conditions occurring in GIFT-Eval and combined them using their frequencies across the 97 tasks.
Encoder forecasts were produced directly in one pass, whereas decoder forecasts used cached patch-recursive generation.

%%%%%%%%%%%%%%%%%%%%%%%%%%%%%%%%%%%%%%%%%%%%%%%%%%%%%%%%%%%%

\newpage
\section*{NeurIPS Paper Checklist}

%%% BEGIN INSTRUCTIONS %%%
The checklist is designed to encourage best practices for responsible machine learning research, addressing issues of reproducibility, transparency, research ethics, and societal impact. Do not remove the checklist: {\bf The papers not including the checklist will be desk rejected.} The checklist should follow the references and follow the (optional) supplemental material.  The checklist does NOT count towards the page
limit. 

Please read the checklist guidelines carefully for information on how to answer these questions. For each question in the checklist:
\begin{itemize}
    \item You should answer \answerYes{}, \answerNo{}, or \answerNA{}.
    \item \answerNA{} means either that the question is Not Applicable for that particular paper or the relevant information is Not Available.
    \item Please provide a short (1--2 sentence) justification right after your answer (even for \answerNA). 
   % \item {\bf The papers not including the checklist will be desk rejected.}
\end{itemize}

{\bf The checklist answers are an integral part of your paper submission.} They are visible to the reviewers, area chairs, senior area chairs, and ethics reviewers. You will also be asked to include it (after eventual revisions) with the final version of your paper, and its final version will be published with the paper.

The reviewers of your paper will be asked to use the checklist as one of the factors in their evaluation. While \answerYes{} is generally preferable to \answerNo{}, it is perfectly acceptable to answer \answerNo{} provided a proper justification is given (e.g., error bars are not reported because it would be too computationally expensive'' or ``we were unable to find the license for the dataset we used''). In general, answering \answerNo{} or \answerNA{} is not grounds for rejection. While the questions are phrased in a binary way, we acknowledge that the true answer is often more nuanced, so please just use your best judgment and write a justification to elaborate. All supporting evidence can appear either in the main paper or the supplemental material, provided in appendix. If you answer \answerYes{} to a question, in the justification please point to the section(s) where related material for the question can be found.

IMPORTANT, please:
\begin{itemize}
    \item {\bf Delete this instruction block, but keep the section heading ``NeurIPS Paper Checklist"},
    \item  {\bf Keep the checklist subsection headings, questions/answers and guidelines below.}
    \item {\bf Do not modify the questions and only use the provided macros for your answers}.
\end{itemize}

%%% END INSTRUCTIONS %%%

\begin{enumerate}

\item {\bf Claims}
    \item[] Question: Do the main claims made in the abstract and introduction accurately reflect the paper's contributions and scope?
    \item[] Answer: \answerYes{} % Replace by \answerYes{}, \answerNo{}, or \answerNA{}.
    \item[] Justification: The claims made match the experimental results.

\item {\bf Limitations}
    \item[] Question: Does the paper discuss the limitations of the work performed by the authors?
    \item[] Answer: \answerYes{} % Replace by \answerYes{}, \answerNo{}, or \answerNA{}.
    \item[] Justification: The limitations are explained in \cref{sec:conclusion}

\item {\bf Theory assumptions and proofs}
    \item[] Question: For each theoretical result, does the paper provide the full set of assumptions and a complete (and correct) proof?
    \item[] Answer: \answerNA{} % Replace by \answerYes{}, \answerNo{}, or \answerNA{}.
    \item[] Justification: There are no theoretical results.

    \item {\bf Experimental result reproducibility}
    \item[] Question: Does the paper fully disclose all the information needed to reproduce the main experimental results of the paper to the extent that it affects the main claims and/or conclusions of the paper (regardless of whether the code and data are provided or not)?
    \item[] Answer: \answerYes{} % Replace by \answerYes{}, \answerNo{}, or \answerNA{}.
    \item[] Justification: Full experimental details are provided in the appendix.

\item {\bf Open access to data and code}
    \item[] Question: Does the paper provide open access to the data and code, with sufficient instructions to faithfully reproduce the main experimental results, as described in supplemental material?
    \item[] Answer: \answerYes{} % Replace by \answerYes{}, \answerNo{}, or \answerNA{}.
    \item[] Justification: The code will be open sourced and published alongside the paper. Instructions will be provided to run each of the experiment.

\item {\bf Experimental setting/details}
    \item[] Question: Does the paper specify all the training and test details (e.g., data splits, hyperparameters, how they were chosen, type of optimizer) necessary to understand the results?
    \item[] Answer: \answerYes{} % Replace by \answerYes{}, \answerNo{}, or \answerNA{}.
    \item[] Justification: Full experiment details are provided in the appendix.

\item {\bf Experiment statistical significance}
    \item[] Question: Does the paper report error bars suitably and correctly defined or other appropriate information about the statistical significance of the experiments?
    \item[] Answer: \answerYes{} % Replace by \answerYes{}, \answerNo{}, or \answerNA{}.
    \item[] Justification: Experiments are ran over multiple seeds and error bars are shown in several plots.

\item {\bf Experiments compute resources}
    \item[] Question: For each experiment, does the paper provide sufficient information on the computer resources (type of compute workers, memory, time of execution) needed to reproduce the experiments?
    \item[] Answer: \answerYes{} % Replace by \answerYes{}, \answerNo{}, or \answerNA{}.
    \item[] Justification: The infrastructure is specified in the paper.
    
\item {\bf Code of ethics}
    \item[] Question: Does the research conducted in the paper conform, in every respect, with the NeurIPS Code of Ethics \url{https://neurips.cc/public/EthicsGuidelines}?
    \item[] Answer: \answerYes{} % Replace by \answerYes{}, \answerNo{}, or \answerNA{}.
    \item[] Justification: Nothing violates the NeurIPS Code of Ethics.

\item {\bf Broader impacts}
    \item[] Question: Does the paper discuss both potential positive societal impacts and negative societal impacts of the work performed?
    \item[] Answer: \answerNA{} % Replace by \answerYes{}, \answerNo{}, or \answerNA{}.
    \item[] Justification: No societal impact.
    
\item {\bf Safeguards}
    \item[] Question: Does the paper describe safeguards that have been put in place for responsible release of data or models that have a high risk for misuse (e.g., pre-trained language models, image generators, or scraped datasets)?
    \item[] Answer: \answerNA{} % Replace by \answerYes{}, \answerNo{}, or \answerNA{}.
    \item[] Justification: The paper poses no such risks.

\item {\bf Licenses for existing assets}
    \item[] Question: Are the creators or original owners of assets (e.g., code, data, models), used in the paper, properly credited and are the license and terms of use explicitly mentioned and properly respected?
    \item[] Answer: \answerYes{} % Replace by \answerYes{}, \answerNo{}, or \answerNA{}.
    \item[] Justification: Part of the dataset and some of the models existed and were introduced prior to this work. They are cited.

\item {\bf New assets}
    \item[] Question: Are new assets introduced in the paper well documented and is the documentation provided alongside the assets?
    \item[] Answer: \answerYes{} % Replace by \answerYes{}, \answerNo{}, or \answerNA{}.
    \item[] Justification: We will release, with this paper, code for the model, the data mixture and the trainer we used to run the experiments. It will be available and open source.

\item {\bf Crowdsourcing and research with human subjects}
    \item[] Question: For crowdsourcing experiments and research with human subjects, does the paper include the full text of instructions given to participants and screenshots, if applicable, as well as details about compensation (if any)? 
    \item[] Answer: \answerNA{} % Replace by \answerYes{}, \answerNo{}, or \answerNA{}.
    \item[] Justification: The paper does not involve crowdsourcing nor research with human subjects.

\item {\bf Institutional review board (IRB) approvals or equivalent for research with human subjects}
    \item[] Question: Does the paper describe potential risks incurred by study participants, whether such risks were disclosed to the subjects, and whether Institutional Review Board (IRB) approvals (or an equivalent approval/review based on the requirements of your country or institution) were obtained?
    \item[] Answer: \answerNA{} % Replace by \answerYes{}, \answerNo{}, or \answerNA{}.
    \item[] Justification: The paper does not involve crowdsourcing nor research with human subjects.

\item {\bf Declaration of LLM usage}
    \item[] Question: Does the paper describe the usage of LLMs if it is an important, original, or non-standard component of the core methods in this research? Note that if the LLM is used only for writing, editing, or formatting purposes and does \emph{not} impact the core methodology, scientific rigor, or originality of the research, declaration is not required.
    %this research? 
    \item[] Answer: \answerNA{} % Replace by \answerYes{}, \answerNo{}, or \answerNA{}.
    \item[] Justification: The core method development in this research does not involve LLMs as any important, original, or non-standard components.

\end{enumerate}

\end{document}